\documentclass[runningheads]{llncs}
\usepackage{graphicx,wrapfig}
\usepackage{algorithm, amsmath}
\usepackage{algpseudocode}
\usepackage{enumitem}
\usepackage{xcolor}
\usepackage{caption}
\usepackage{subcaption}

\begin{document}
\title{Learning Semi-Structured Representations of Radiology Reports}
%
%
\author{Tamara Katic\inst{1}
\and
Martin Pavlovski\inst{1}
\and
Danijela Sekulic\inst{2} \and 
Slobodan Vucetic\inst{1}}
\authorrunning{T. Katic, M. Pavlovski, D. Sekulic, S. Vucetic}
%
\institute{Temple University, Philadelphia, PA, USA \and University Clinical Center of Serbia, Belgrade, Serbia \\
\email{\{tamara.katic,martin.pavlovski,slobodan.vucetic\}@temple.edu,dacasekulic@gmail.com}}

\maketitle              
\vspace{-10mm}
\begin{abstract}
Beyond their primary diagnostic purpose, radiology reports have been an invaluable source of information in medical research. Given a corpus of radiology reports, researchers are often interested in identifying a subset of reports describing a particular medical finding. Because the space of medical findings in radiology reports is vast and potentially unlimited, recent studies proposed mapping free-text statements in radiology reports to semi-structured strings of terms taken from a limited vocabulary. This paper aims to present an approach for the automatic generation of semi-structured representations of radiology reports. The approach consists of matching sentences from radiology reports to manually created semi-structured representations, followed by learning a sequence-to-sequence neural model that maps matched sentences to their semi-structured representations. We evaluated the proposed approach on the OpenI corpus of manually annotated chest x-ray radiology reports. The results indicate that the proposed approach is superior to several baselines, both in terms of (1) quantitative measures such as BLEU, ROUGE, and METEOR and (2) qualitative judgment of a radiologist. The results also demonstrate that the trained model produces reasonable semi-structured representations on an out-of-sample corpus of chest x-ray radiology reports from a different medical provider.


\vspace{-2mm}

\keywords{automatic document annotation \and semi-structured representations \and radiology reports \and chest x-ray \and sequence-to-sequence.}\vspace{-4.5mm}
\end{abstract}
%
%
\section{Introduction}
\begin{wrapfigure}{r}{0.5\textwidth}
    \centering
    \vspace{-0.8mm}
    \includegraphics[width=0.48\textwidth]{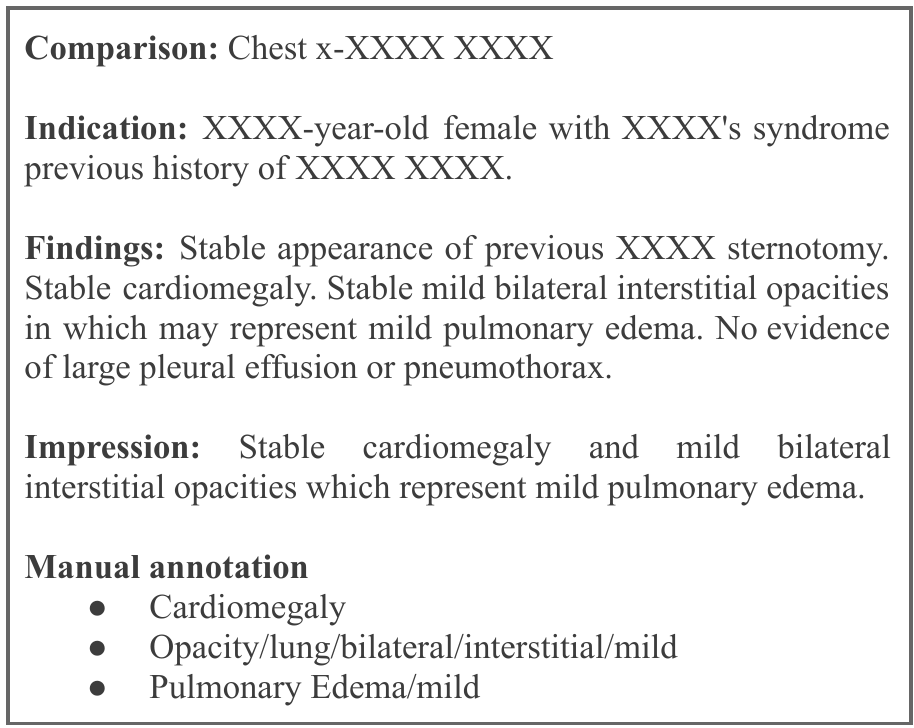}
    \vspace{-3.5mm}
    \caption{An example of \cite{openilink} radiology report with manual annotations.}
    \label{fig:report_example}
    \vspace{-8mm}
\end{wrapfigure}

\vspace*{-2mm}
\label{sec:intro} There has been an increasing interest in automatically extracting information from free-text portions of medical reports~\cite{enayati2021visualization}. Radiology reports provide significant information and thus are one of the most broadly researched free-text report types for information extraction. An example of a radiology report is shown in Figure~\ref{fig:report_example}. It typically consists of the following sections: \textit{Comparison}, \textit{Indication}, \textit{Findings}, and \textit{Impression}. \textit{Findings} is the key section that provides a detailed description of the main observations radiologists made by studying a radiology image. \textit{Impression} is a brief summary of the most significant observations.

Figure~\ref{fig:report_example} also shows  \textit{Manual annotation}, which is an outcome of the OpenI~\cite{demner2016preparing} project, where the objective was to use a limited vocabulary to explain all significant observations from \textit{Findings} and \textit{Impression} as a semi-structured text. The hypothesis was that such semi-structured text would enable training an accurate sequence-to-sequence model that converts free-text radiology reports into more
easily queried representations. Each bullet point from Figure~\ref{fig:report_example} shows one annotation, and each annotation may contain between one and eight terms. The terms are taken from Medical Subject Headings (MeSH)\footnote{\url{https://www.nlm.nih.gov/mesh/meshhome.html}} and RadLex\footnote{\url{http://radlex.org}} vocabularies. It took OpenI radiology experts between 10 and 20 minutes to annotate each radiology report~\cite{demner2015annotation}. Thus, it is not feasible to expect radiologists to provide semi-structured summaries as part of their regular practice.

\vspace{1mm}
In this paper, we propose a novel approach to automatically generate semi-structured representations of radiology reports. The first step of the approach is to match each manual annotation with its corresponding sentence from the \textit{Findings} or \textit{Impression} sections. The second step is to train a sequence-to-sequence neural network that maps each  sentence to its semi-structured representation. If successful, the trained neural network can be applied on a large corpus of unlabeled radiology reports and enable information extraction. This could be very useful for operational and research tasks, such as the identification of patients with a particular diagnosis.

The main contributions of this paper are as follows:
\begin{itemize}[noitemsep,topsep=2pt,leftmargin=*]
    \item[•] We propose a procedure that automatically matches OpenI annotations with sentences from radiology reports (see Section~\ref{sec:SAL}).
    \item[•] We train a sequence-to-sequence neural network for automatic annotation of radiology reports (see Section~\ref{sec:sentence_model}).
    \item[•] We compare the proposed approach with several baselines (see Section~\ref{sec:results}).
    \item[•] We demonstrate that the semi-structured representations generated by our approach are accurate both on an in-sample and out-of-sample corpus of radiology reports (see Section~\ref{sec:radiologist_results}).
\end{itemize}
\section{Related Work}
\label{sec:related}
\vspace*{-2mm}
\textbf{Annotations from Radiology Reports.} Most prior works have attempted to automatically detect the presence of key findings (e.g., disease diagnoses) from radiology reports \cite{hassanpour2017characterization,irvin2019chexpert,peng2018negbio,wang2017chestx,wang2018tienet,smit2020chexbert} without much contextual information such as location and severity of diseases.
On the other hand, Shin et al.~\cite{shin2016learning} trained a model that learned from both text and images to produce annotations that include both diseases and their contexts.
They used only reports from the OpenI dataset that contain the 17 most frequent MeSH terms, which resulted in approximately 40\% of all OpenI reports. Conversely, this paper utilizes the original OpenI corpus with manual annotations covered by 101 MeSH terms and 76 RadLex terms.
\\
\textbf{Mapping Radiology Terms to a Pre-defined Vocabulary.} Datta et al.~\cite{datta2020radlex} created and normalized a manually-annotated corpus of radiology reports. They mapped findings, medical devices, and procedures to the publicly available radiology lexicon-RadLex. However, some terms could not be mapped due to the lack of exact matches in RadLex.
Demner-Fushman et al.~\cite{demner2015annotation} also found no exact matches of several terms from their reports to the RadLex, UMLS\footnote{\url{https://www.nlm.nih.gov/research/umls/index.html}}, and MeSH lexicons. Due to the absence of publicly available annotated datasets, most of these works require human effort to annotate several thousands of reports.\\
\textbf{Summarization of Radiology Findings.} Several text summarization methods~\cite{nallapati2016abstractive,see2017get,paulus2017deep,chen2018fast} were used to summarize clinical notes~\cite{zhang2018learning,macavaney2019ontology,liu2019clinically}.
Summarizing a report's \textit{Findings} paragraph into an \textit{Impression} paragraph was explored in~\cite{zhang2018learning,sotudeh2020attend}. On the other hand, in this paper, we study the summarization of both of these paragraphs into shorter sequences of annotation terms.\vspace*{-2mm}
\section{Dataset}
\vspace*{-2mm}
\label{sec:dataset} 
We use the publicly available OpenI dataset\footnote{\url{https://openi.nlm.nih.gov/services\#searchAPIUsingGET}}~\cite{demner2016preparing} collected by Indiana Network for Patient Care. The dataset consists of 3,995 annotated radiology reports, which are associated with 7,470 radiology images. To the best of our knowledge, there are no large corpora of annotated with contextual information (particularly not by a domain expert) radiology reports available in the NLP community. 
That being said, we are using a small corpus with high-quality annotations. Each of the 3,995 reports is manually annotated using a limited number of MeSH and RadLex terms. The terms can be classified into five categories: \textit{diseases}, \textit{anatomy}, \textit{objects}, \textit{signs}, and \textit{attributes}~\cite{demner2015annotation}. Each annotation consists of a sequence of several terms, separated by a special symbol (``/'') and represents a single significant radiology finding. The first term, or \textit{heading}, are from the \textit{disease}, \textit{anatomy}, \textit{object}, or \textit{sign} category. The following terms, or \textit{subheadings}, are qualifiers and belong to the \textit{anatomy} or \textit{attributes} category. An example of one report with manual annotations under the \textit{Manual annotation} section is shown in Figure~\ref{fig:report_example}.
\vspace*{-1.8mm}
\section{The Proposed Approach}
\label{sec:model}
\vspace*{-2mm}
OpenI dataset provides a set of semi-structured annotations for each radiology report. The baseline approach used in our experiments for automatic generation of semi-structured annotations takes a whole radiology report (treating it as a single paragraph) as an input and produces a sequence of annotations as an output. Our hypothesis is that it would be beneficial to (1) first match each annotation in a report with its corresponding sentence; and then to (2) use the matched sentence-annotation pairs to train a model capable of automatically generating sentence annotations.
Thus, we propose an approach that generates annotations at a sentence level rather than a paragraph/report level. Our approach consists of two components: (1) a sentence-annotation matching procedure and (2) a sequence-to-sequence neural network for sentence annotation generation.
\vspace*{-2mm}
\subsection{Sentence-Annotation Matching}

\vspace*{-2mm}
\begin{figure}[!t]
    \centering
    \includegraphics[width=0.6\textwidth]{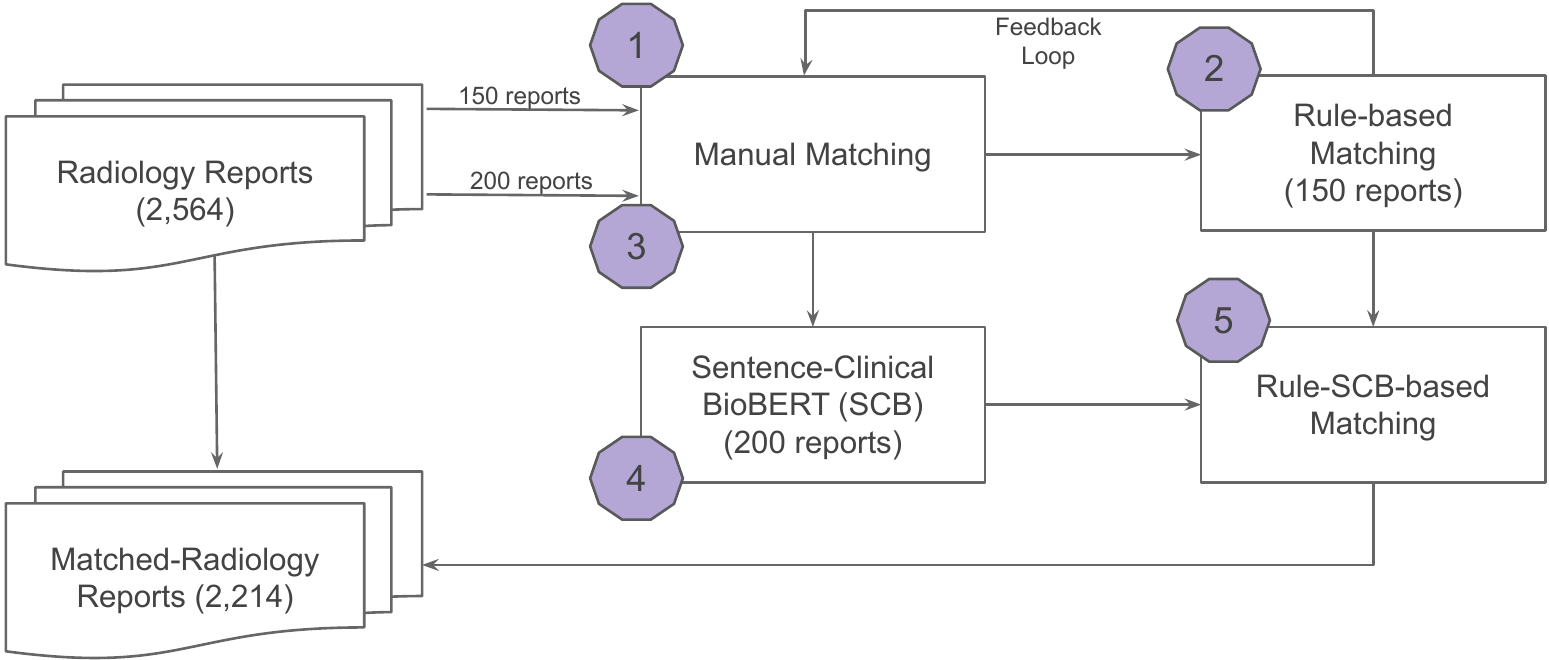}
    \caption{Steps of the proposed procedure: (1) manual matching of 150 reports, (2) building a rule-based algorithm to perform weak matching, (3) manual matching of additional 200 reports for additional performance improvement, (4) training of a Sentence Transformer (Sentence-Clinical BioBERT) model to find the semantic similarity of sentence-annotation pairs, (5) combining steps (2) and (4) to find sentence-annotation pairs in the rest of 2,214 reports.}
    \label{fig:labeler}
    \vspace{-4mm}
\end{figure}

\label{sec:SAL}
The process of matching sentences with annotations starts with the manual matching of a limited number of OpenI reports. It is followed by the creation of manually constructed matching rules. Finally, a sentence-annotation matching algorithm is trained using a pre-trained neural language model. Our proposed approach is described in detail below and is illustrated in Figure~\ref{fig:labeler}.\\
\textbf{Manual Matching.}
We randomly selected 150 OpenI reports and manually matched their annotations with sentences from the Findings and Impression sections by consulting the annotation rules described by Demner-Fushman et al. ~\cite{demner2016preparing}. We used both sections because Gershanik et al.~\cite{gershanik2011critical} showed that the Impression section is not always a summary of the Finding section. Both sections are collectively referred to as \textit{findings} in this paper. The \textit{findings} paragraph from a radiology report is represented as a sequence of sentences $S=\{s_1,s_2,\dots,s_N\}$ and the set of all possible manual annotations is denoted as $A=\{a_1,a_2,\dots,a_M\}$, where annotation $a_j$ is represented as a sequence of terms, for each $j=1,\dots,M$. Note that $N$ does not need to equal $M$. For each report, our goal is to find an $s_i$ that corresponds to an $a_j$. Based on our experience, we have observed that each of almost all annotations can be matched with a single sentence, while it is possible that some complex sentences end up being annotated with multiple annotations.
On the other hand, many sentences may not be matched with any annotation. Often, those sentences describe negative findings, where radiologists explain that they are not observing a concerning pathology.\\
\phantom{iiiii}By reading sentences from our reports, a radiologist mapped the annotations' terms to their synonyms in the sentences. The final list of term-synonym pairs is defined as a dictionary $D$ (e.g., \textit{scarring: cicatrix, 
pulmonary disease, chronic obstructive: copd}). The radiologist manually matched sentences and annotations of 150 reports, which served as ground truth to our rule-based algorithm.\\
\textbf{Rule-based Matching.}
To automatically match sentences and annotations, we developed a rule-based algorithm. Inputs to our rule-based matching system are sentences $S$ and annotations $A$ from a radiology report and dictionary $D$ of term-synonym pairs. For each $a_j \in A$, the rule-based algorithm attempts to find the sentence $s_i \in S$ that best corresponds to $a_j$. As explained before, each annotation is composed of a single term called \textit{heading}, followed by a list of terms called \textit{subheadings}. The first step is to extract the \textit{heading} and create a list of candidate words (\textit{h\_words}) based on both syntactic and semantic similarities. The \textit{h\_words} list consists of the $n$-grams of the \textit{heading}, stemma~\cite{porter1980algorithm}, synonyms from dictionary $D$, and the $k$ (= 5) most similar words determined by a trained FastText model (see Section~\ref{sec:experimental_setup}). The second step is to find $s_i$ with the highest number of matches from the \textit{h\_words}. If the algorithm matches only one sentence $s_i$ with the annotation $a_j$ or the annotation $a_j$ has only the \textit{heading}, the algorithm returns $s_i$. Otherwise, the algorithm proceeds by matching with the \textit{subheadings}. Similarly to the previous step, the algorithm creates a list of candidate words (\textit{sh\_words}) using word-level $n$-grams and synonyms of \textit{subheadings} from dictionary $D$. Then, the sentence with the highest number of matched \textit{sh\_words} is selected.\\
\textbf{Sentence-Clinical BioBERT (SCB).}
To improve the performance of the rule-based algorithm, we asked the radiologist to manually match sentences with annotations from additional 200 radiology reports in the same manner as with the previous (manually matched) 150 reports. For further details on the choice of the 150/200 split, refer to Section~\ref{sec:experimental_setup}. The resulting matched pairs are then used to train a Sentence Transformer (Sentence-BERT) model~\cite{reimers2019sentence} to find sentence-annotation pairs with the highest similarity scores. \\
\phantom{iiiii}We build the Sentence-BERT architecture from scratch by defining its layers individually. First, we define the embedding layer with the Clinical BioBERT model~\cite{alsentzer2019publicly} to generate word embeddings. Second, we create a mean pooling layer that performs pooling on the word embeddings and returns the fixed size sentence embedding. Thus, regardless of how long the input text is, the output vector is a fixed 768-dimensional. Based on the aforementioned two layers, we build a SCB model. To generate sentence and annotation embeddings, we pass each sentence and annotation through our network separately. The similarity scores calculated for each pair of sentence and annotation embeddings are then thresholded. If a pair's similarity score is above the threshold, the pair is labeled with ``one'', otherwise it is labeled with ``zero''. \\
\phantom{iiiii}Input to the SCB model are the sentence-annotation pairs, and the output are the binary labels. In this experiment, we preprocessed the annotations by removing slashes in order to convert them into a sentence form. Each manually matched sentence-annotation pair from the 200 reports is labeled with ``one''. To train SCB, we need to create dissimilar pairs of sentence-annotation. We randomly select two sentence-annotation pairs per report, unrelated to each other, and label them with ``zero''. A few examples of similar and dissimilar pairs are presented in Table~\ref{tab:sent_annot_pairs}.\\
\textbf{Rule-SCB-based Matching.}
The trained SCB model is incorporated as the last rule in the rule-based matching algorithm (see ``Rule-based Matching'' in Section~\ref{sec:SAL}). Finally, this algorithm is applied to the remaining 2,214 reports to \\
\clearpage
\begin{table}[!t]
\centering
\caption{Several examples of sentence-annotation pairs.}
\resizebox{\textwidth}{!}{%
\begin{tabular}{|l|l|c|}
\hline
\textbf{Sentence} & \textbf{Annotation} & \textbf{Label}\\
\hline
There is no pneumothorax or pleural effusion. & Opacity lung base left mild. & 0 \\
Calcified hilar lymph. & Calcinosis lung hilum lymph nodes. & 1 \\
No acute disease. & Lung hyperdistention. & 0 \\
Low lung volumes. & Lung hypoinflation. & 1 \\ \hline
\end{tabular}}
\label{tab:sent_annot_pairs}
\end{table}

\begin{algorithm}[!t]
	\caption{ Match Sentences with Annotations }
	\label{alg:SAL}
	\begin{algorithmic}
	\State \hspace{-0.5cm} \textbf{procedure} Rule\_SCB($S, A, D$)
        \For{i=1...$|A|$}
    		\State $h = heading(A[i])$
            \State $h\_words$ = $candidate\_words(h)$
    		\State find all $S_m \subset S$ with highest \# of $h\char`_words$ 
    		\If{$|S_m|$ = 1}
    		\State $s = S_m$
    		\ElsIf{length($A[i]$) = 1}
    		\State $s = S_m[1]$
    		\ElsIf{$|S_m|$ $>$ 1}
    		\State $sh$ = $subheadings(A[i])$
    		\State $sh\char`_words$ = $candidate\_words(sh)$
    		\State find $s \in S_m$ with highest \# of $sh\char`_words$
    		\Else
            \State find $s \in S_m$ to $A[i]$ with SCB
    		\EndIf
    		\State label $s$ with $A[i]$
    	\EndFor
	\end{algorithmic}
\end{algorithm}
\phantom{i}
\vspace{-7mm}
\\
match their sentences and annotations. The described approach is outlined in Figure~\ref{fig:labeler} and its pseudocode is given in Algorithm~\ref{alg:SAL}.

\subsection{Sentence Annotation Generation (SAG)}
\label{sec:sentence_model}
This section describes a sequence-to-sequence model for automatic generation of sentence annotations, abbreviated as SAG-Seq2Seq.\\
\textbf{Input and output.} Input to SAG-Seq2Seq is a sentence, and corresponding one or more annotations are at the output; however, if the sentence lacks an annotation\footnote{These sentences are without any abnormalities.}, the output is set to an end-of-sequence token (``.''). In reports with multiple annotations, the order of the annotations follows the order of sentences, followed by an end-of-sequence token.\\
\textbf{Task.} Our task has similarities with typical sequence-to-sequence tasks such as text translation and text summarization. It can be thought of as (1) a task of translating an input sentence (being a sequence of words) into an annotation (sequence of terms), or (2) a text summarization task in which the input sentence is essentially ``summarized'' into a shorter sequence of annotation terms. To generate annotations from the input sentence, we use models based on a well-known encoder-decoder architecture~\cite{sutskever2014sequence} and rely on an attention mechanism~\cite{bahdanau2014neural}.\\
\textbf{Learning.} Inspired by the work of Zhang et al. ~\cite{zhang2018learning}, SAG-Seq2Seq incorporates a pointer-generator network~\cite{see2017get}. We exclude the coverage mechanism from this network as it led to lower performance. The architecture of this network is further described.  A sequence of words $w=\{w_1,w_2,...,w_N\}$ from each sentence is fed into an encoder (two-layer BiLSTM), resulting in a sequence of hidden states $h=BiLSTM(w)$. After encoding the entire input sequence, the output sequence is generated in a step-by-step manner using a separate decoder (single-layer LSTM). The decoder calculates the current state $s_t$ using the previously generated token and the previous decoder state $s_{t-1}$. To achieve better decoding, the attention mechanism is used to assist the decoder with where to look in order to produce the next word. The attention weights are calculated using a softmax function over the decoder state $s_t$ given an input hidden state $h_i$. Then, the attention weights are employed to compute a context vector. Lastly, the context vector and decoder state $s_t$ are used to either copy the next token from the input sequence or generate it from the vocabulary.\\
\textbf{Inference.} The trained SAG-Seq2Seq is applied to each sentence from a given report's \textit{findings} paragraph. 
To obtain a unique set of annotations for the report, we take the union of all generated sentence annotations.

\begin{figure}[!t]
    \centering
    \includegraphics[width=0.5\textwidth]{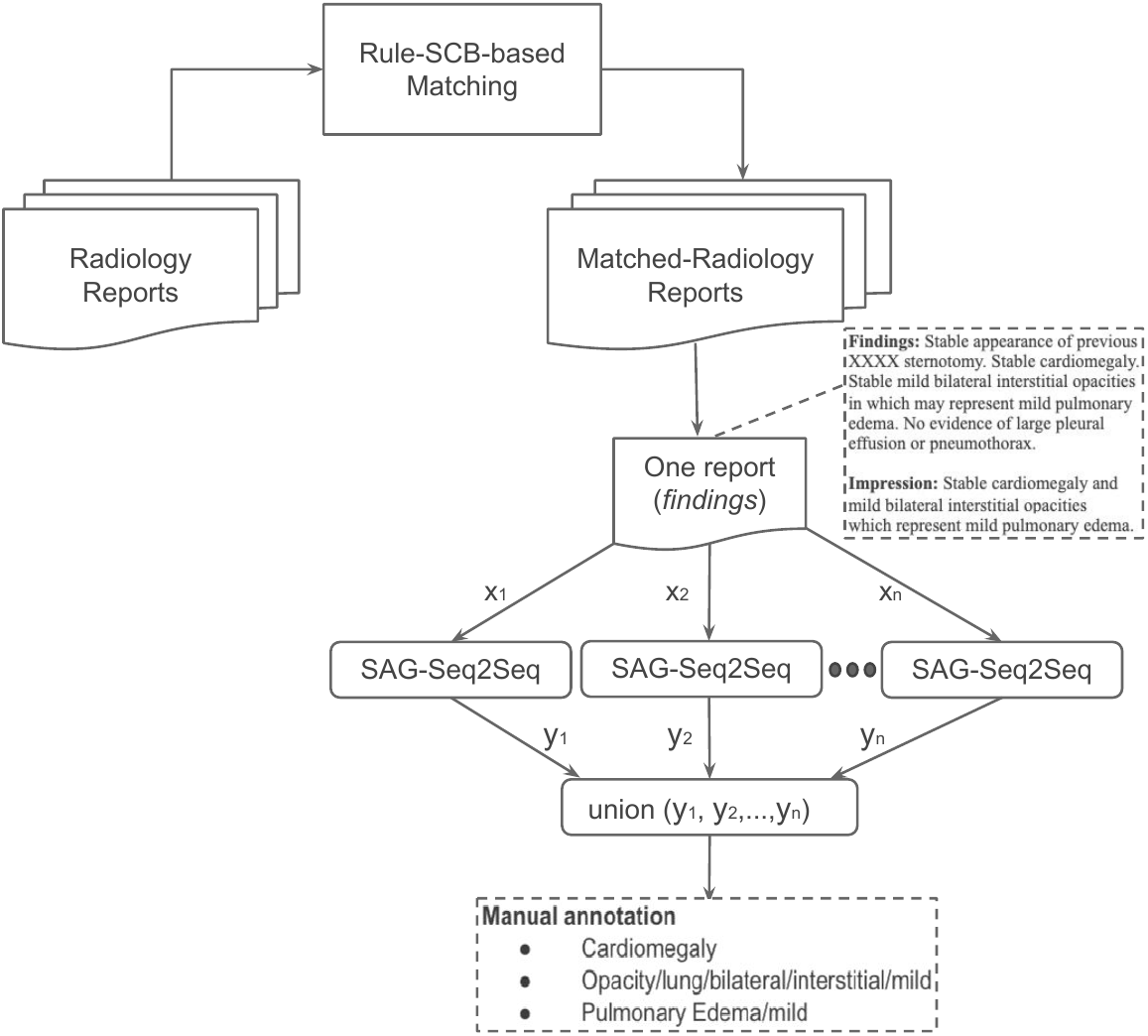}
    \vspace{-3mm}
    \caption{Overall architecture of our approach for automatic generation of annotations.}
    \label{fig:overall_architecture}
    \vspace{-6mm}
\end{figure}

The overall architecture of our proposed approach is illustrated in Figure~\ref{fig:overall_architecture}. 

\vspace*{-2mm}
\section{Experiments}
\label{sec:experiments}
\vspace*{-2mm}
\subsection{Data Selection}
\label{sec:data_selection}
\vspace{-0.4mm}
The unbalanced ratio of diagnosed to healthy cases is a common challenge in medical analysis since the former are less frequent in the general population. We filtered the dataset with 3,955 reports by excluding 1,391 reports (around 35\%) related to healthy cases (marked as \textit{``normal''} in the \textit{Manual annotation} section).  Even though we deleted those reports the remaining still contained sentences
\begin{table}[!t]
\parbox{0.43\textwidth}{
\vspace{5mm}
\normalsize
without disease terms ($\sim$68\%). Additionally, we removed the ``XXXX'' patterns that represent words or phrases that were initially replaced by the de-identification algorithm~\cite{demner2016preparing}, digits, punctuations and extra spaces. Table~\ref{tab:corpus_stats} shows more detailed statistics of the preprocessed corpus. We randomly split the dataset into 80\% for training, 10\% for validation, and 10\% for testing.
}
\hspace{2mm}
\parbox{0.5\textwidth}{
\centering
\caption{Statistics of our final corpus.}
\begin{tabular}{|l|c|}
\hline
\# reports & 2,564 \\
\# sentences & 16,400 \\
\# annotations & 6,907 \\
\hline
\# sentences without annotations & 11,153 \\
\# sentences with annotations & 5,247 \\
\# sentences with only one annotation & 4,095 \\
\# sentences with several annotations & 1,152 \\
\hline
average \# of words in sentences & 6.65 \\
average \# of words in annotations & 1.40 \\ \hline
\end{tabular}
\label{tab:corpus_stats}
}
\vspace*{-3mm}
\end{table}
\subsection{Baseline Models}
\label{sec:baselines}
\vspace*{-1mm}
\textbf{Paragraph-level Models.} Our model is compared against the two well-known models for text summarization. The input to these models is the \textit{findings} paragraph. Thus, we refer to them as paragraph-level models. Same as for SAG-Seq2Seq, we used the 100-dimensional word vectors generated by FastText (see Section~\ref{sec:experimental_setup}) to initialize the word embeddings of the following baseline models:
\begin{itemize}[noitemsep,topsep=2pt,leftmargin=*]
    \item[•] Paragraph-level Pointer-Generator Model (\textbf{PL-PG}). We use a sequence-to-sequence attention summarization model referred to as Pointer-Generator (PG)~\cite{see2017get}. Note that the coverage mechanism was excluded from this baseline since it did not improve the quality of generated annotations in our task.
    \item[•] Paragraph-level Pointer-Generator Model with Background information (\textbf{PL-PG+Background}). An extension of the Pointer-Generator network~\cite{zhang2018learning} which utilizes a separate attentional encoder that encodes the background information from the radiology reports' \textit{Indication} sections and uses it to guide the decoding process of the overall model.
\end{itemize}
\textbf{Sentence-level Models.} SAG-Seq2Seq was also compared to four transformer encoder-decoder models (whose inputs are sentences, same as for SAG-Seq2Seq):
Bidirectional and Auto-Regressive Transformers (\textbf{BART}) \cite{lewis2019bart}, MAsked Sequence to Sequence pre-training (\textbf{MASS})~\cite{song2019mass}, \textbf{MarianNMT} (an efficient Neural Machine Translation framework)~\cite{junczys2018marian}, and \textbf{Transformer+PG} \cite{enarvi2020generating}. For the first three models, we used their variants pre-trained on clinical text (using the Clinical BioBert tokenizer) and fine-tuned them on our sentence annotation generation task. Furthermore, using the fourth model, we explored the performance of replacing our RNN-based with a Transformer-based encoder-decoder model with a pointing mechanism. As opposed to the previous transformer-based baselines, we trained this model from scratch.
\vspace*{-2mm}
\subsection{Experimental Setup}
\label{sec:experimental_setup}
\textbf{Evaluation Metrics.} Sentence-annotation matching is evaluated using accuracy by comparing the predicted labels against the ground truth. All models that are trained to generate annotations are evaluated using BLEU~\cite{papineni2002bleu}, METEOR~\cite{banerjee2005meteor}, and ROUGE-L $\mathrm{F}_1$~\cite{lin2004rouge}. The BLEU-N scores are evaluated for cases with at least N annotation terms. Here, we compute BLUE-1 to BLEU-4.\\
\textbf{Word Vectors.} Due to the small size of the OpenI dataset, the MIMIC-CXR dataset\footnote{We extracted the reports that have either the \textit{Findings} or \textit{Impression} section.}~\cite{johnson2019mimic} is additionally used to produce word vectors. The corpus has 218,870 reports (216,306 reports from MIMIC-CXR and 2,564 reports from OpenI). First, we used the Clinical BioBERT tokenizer to initialize the word vectors. Nevertheless, we have observed that such initialization did not further improve models' performances. Thus, we trained a FastText model instead, using the default hyperparameters in the Gensim implementation\footnote{\url{https://radimrehurek.com/gensim/models/fasttext.html}}.\\ 
\textbf{Threshold for Manual Matching.}
We chose 150 OpenI reports for manual matching as we found that considering any additional reports for model training did not significantly improve the performance. In our case, manually matching 200 reports turned out to be sufficient for training the SCB (see Section~\ref{sec:SAL}) model since a Pearson correlation coefficient of 0.9 was observed between the sentence and annotation representations produced by SCB.\\
\textbf{Training Details.}
SAG-Seq2Seq employed a two-layer BiLSTM encoder with a hidden size of 256 for each direction, and a single-layer LSTM decoder with a hidden size of 512. The parameters of SAG-Seq2Seq were optimized using the Adam optimizer~\cite{kingma2014adam} and a learning rate of 0.001. We used a batch size of 16 and clipped the gradient with a norm of 5. Our annotations are generated using beam search with a beam size of 5.

\begin{table}[!t]
\parbox{.6\linewidth}{
\centering
\caption{Accuracy of sentence-annotation matching methods. ``baseline'' refers to random sentence selection. The remaining methods use both the \textit{heading} and \textit{subheadings} of an annotation to find an appropriate sentence match.
}
\begin{tabular}{|l|c|}
\hline
\textbf{Method} & \textbf{Accuracy} \\
\hline
baseline & 0.1273 \\
$n$-gram matching & 0.4991 \\
$k$-most-similar & 0.5972 \\
$n$-gram matching + term-synonyms & 0.8389 \\
$k$-most-similar + term-synonyms & 0.9288 \\ \hline
Rule-based & 0.9574 \\
Rule-SCB-based & \textbf{0.9895} \\ \hline
\end{tabular}
\label{tab:sent_annot_accuracy}
}
\hspace{1mm}
\parbox{.35\linewidth}{
\centering
\caption{Accuracy obtained when applying each condition from Algorithm~\ref{alg:SAL}, separately.}
\centering
\begin{tabular}{|l|c|}
\hline
\textbf{Condition} & \textbf{Accuracy} \\
\hline
first & 0.6417 \\
second & 0.0857 \\
third & 0.3610 \\
fourth (SCB-based) & 0.8962 \\
\hline
\end{tabular}
\label{tab:sent_annot_accuracy_additional_results}
}
\vspace{-4mm}
\end{table}


\begin{table}[!t]
\centering
\caption{Evaluation of generated annotations on the test set using BLEU, METEOR and ROUGE-L.}
\resizebox{\textwidth}{!}{%
\begin{tabular}{|l|c|c|c|c|c|c|}
\hline
\textbf{Model} & \textbf{BLUE-1} & \textbf{BLEU-2} & \textbf{BLEU-3} & \textbf{BLEU-4} & \textbf{METEOR} & \textbf{ROUGE-L}\\
\hline
PL-PG & 0.7362 & 0.6734 & 0.6200 & 0.5689 & 0.3885 & 0.7469 \\
PL-PG+Background & 0.6907 & 0.6335 & 0.5835 & 0.5359 & 0.3703 & 0.7339 \\\hline
BART & 0.4901 & 0.3784 & 0.3246 & 0.2842 & 0.2203 & 0.4497 \\
MASS & 0.4876 & 0.3813 & 0.3318 & 0.2956 & 0.2214 & 0.4505 \\
MarianNMT & 0.5059 & 0.4044 & 0.3554 & 0.3193 & 0.2438 & 0.4592 \\
Transformer+PG & 0.4966 & 0.3840 & 0.3287 & 0.2874 & 0.2243 & 0.4385 \\ \hline
SAG-Seq2Seq & \textbf{0.7380} & \textbf{0.6986} & \textbf{0.6638} & \textbf{0.6287} & \textbf{0.4111} & \textbf{0.7637} \\ \hline
\end{tabular}}
\label{tab:evaluation_proposed_model}
\vspace{-4mm}
\end{table}
\vspace*{-2mm}
\section{Results and Discussion}
\label{sec:results_and_discussion}
\vspace*{-2mm}
\subsection{Experimental Results}
\label{sec:results}
\vspace*{-1.5mm}
Initially, the 150 randomly selected and manually matched radiology reports from the OpenI dataset (see Section~\ref{sec:SAL}) were used to evaluate the performance of the rule-based algorithm. This algorithm was built in several iterations (see Table~\ref{tab:sent_annot_accuracy}) until results close or same to the ground truth pairs were attained. In most cases, mismatches throughout the iterations occurred due to the lack of rules, the order in which the rules appeared, and the absence of some term-synonym pairs in $D$. By running an ablation study with each if-condition of Algorithm~\ref{alg:SAL}, as shown in Table~\ref{tab:sent_annot_accuracy_additional_results}, we found that better sentence-annotation matching performance is obtained once all conditions are combined as opposed to applying each condition separately. The final results of this experiment are presented in Table~\ref{tab:sent_annot_accuracy}, demonstrating that the rule-SCB-based algorithm attained the largest performance and thus we leverage this rule-based variant within our overall architecture in all of the following experiments.

Next, we have compared SAG-Seq2Seq against the baseline models. The results are presented in Table~\ref{tab:evaluation_proposed_model}. Overall, SAG-Seq2Seq outperforms all baselines across all metrics. PL-PG achieves better performance than the other baselines, contrary to the observation in~\cite{zhang2018learning}. We found that using the separate background encoder (PL-PG+Background) deteriorates the performance. One of the reasons is that the background information per each report is short with only two or three words, while the background information used in~\cite{zhang2018learning} contains sentences with richer patient information. Note that SAG-Seq2Seq has been additionally compared against several transformer encoder-decoder baselines. Although these baselines have achieved state-of-the-art results on various tasks, we observed that SAG-Seq2Seq attained greater performance on our task at hand. Some of the potential reasons why these baselines perform worse may include but are not limited to: term repetition (same terms occurring in both annotations and their corresponding input sentences), and generation of longer annotations (in some cases the model generates annotations with a larger number of terms than those in the ground truth annotations). Moreover, it is worth noting that the observation of transformer-based models performing worse than LSTM-based seq2seq models (such as our proposed model) is supported by~\cite{hu2021word,ezen2020comparison}, where LSTMs outperformed transformer-based models on a sequence-to-sequence task similar to the one considered here.
The main reason behind this could be because LSTMs perform well on small datasets, whereas transformer-based models tend to be more powerful when trained on large amounts of data.

\begin{table}[!t]
\centering
\caption{A test example of annotations generated by SAG-Seq2Seq and the two top competing baselines, as well as the corresponding manual annotations (ground truth). The correctly generated, incorrectly generated and omitted annotation terms are highlighted in blue, red and orange color, respectively. 
}
\begin{tabular}{p{0.25\textwidth}p{0.73\textwidth}}
\hline
Manual annotation & Cardiomegaly/severe, Implanted Medical Device/left, Pulmonary Congestion/mild, Pericardial Effusion \\
\hline\hline
PL-PG & \textcolor{blue}{Cardiomegaly}/\textcolor{red}{mild} \textcolor{orange}{severe}, \textcolor{blue}{Implanted Medical Device/left}, \textcolor{red}{Mild}, \textcolor{blue}{Pulmonary Congestion}\textcolor{orange}{/mild}, \textcolor{orange}{Pericardial Effusion} \\
\hline 
PL-PG+Background & \textcolor{blue}{Cardiomegaly/severe}, \textcolor{blue}{Implanted Medical Device/left}, \textcolor{red}{Mild}, \textcolor{blue}{Pulmonary Congestion}\textcolor{orange}{/mild}, \textcolor{orange}{Pericardial Effusion} \\
\hline
SAG-Seq2Seq & \textcolor{blue}{Cardiomegaly/severe}, \textcolor{blue}{Implanted Medical Device/left}, \textcolor{blue}{Pulmonary Congestion/mild}, \textcolor{blue}{Pericardial Effusion} \\
\hline
\end{tabular}
\label{tab:test_example}
\end{table}

\noindent
\textbf{Case Study.} We visually inspected several examples of sentence annotations generated by SAG-Seq2Seq and two top competing baselines, and compared them to the respective manual annotations (ground truth). 
Table~\ref{tab:test_example} shows a representative test example of annotations generated by the various models.
Both baselines produce incorrect annotations and fail to predict \textit{mild} and \textit{Pericardial Effusion} in this case, whereas our model successfully generates all annotations. Through visual analysis of additional samples from the test set, we observed that our model generated more precise and correct annotations than the baselines.
\begin{figure}[!t]
\centering
\begin{subfigure}[b]{0.95\textwidth}
    \hspace{-3mm}
    \includegraphics[width=0.38\textwidth]{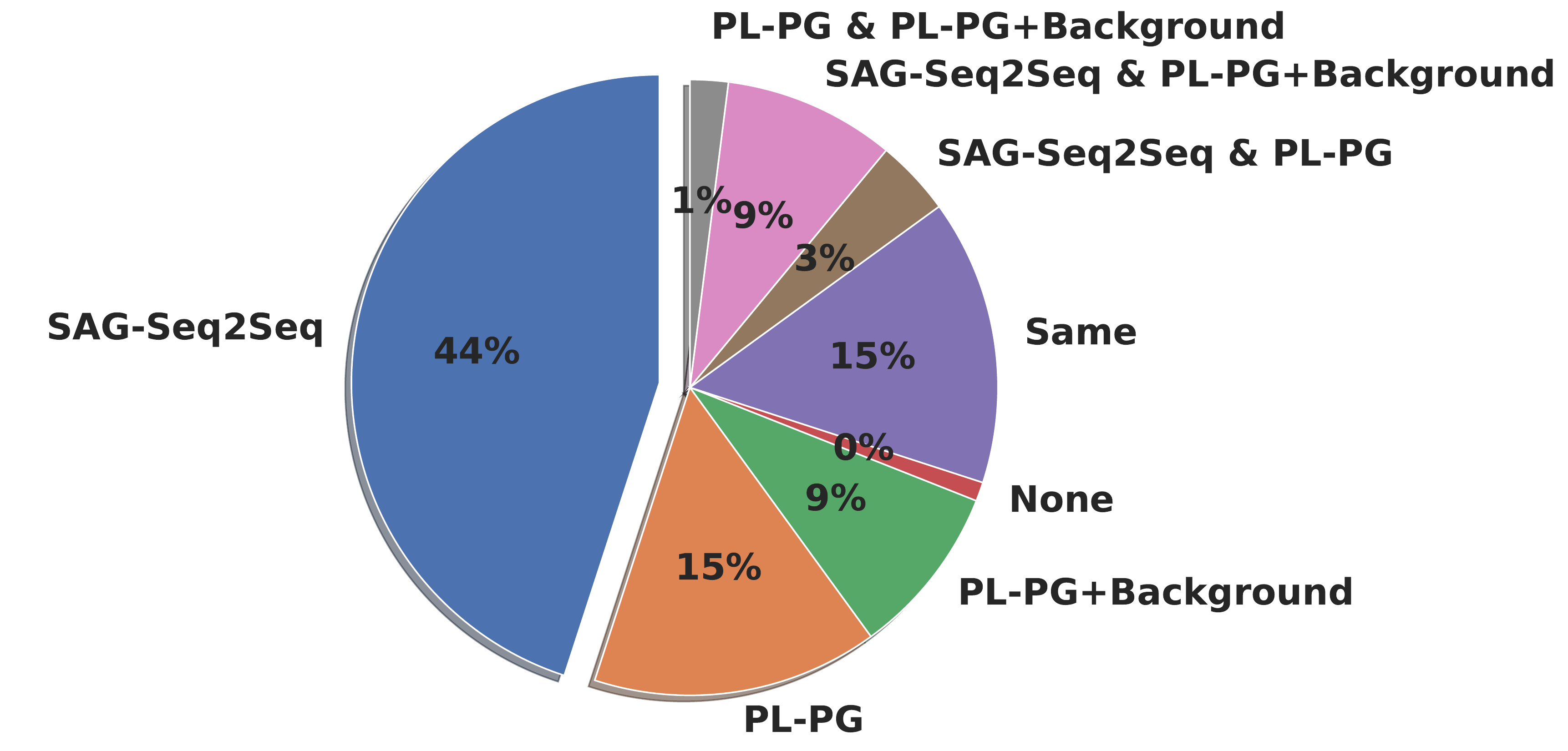}
    \hspace{-1.9mm}
    \includegraphics[width=0.336\textwidth]{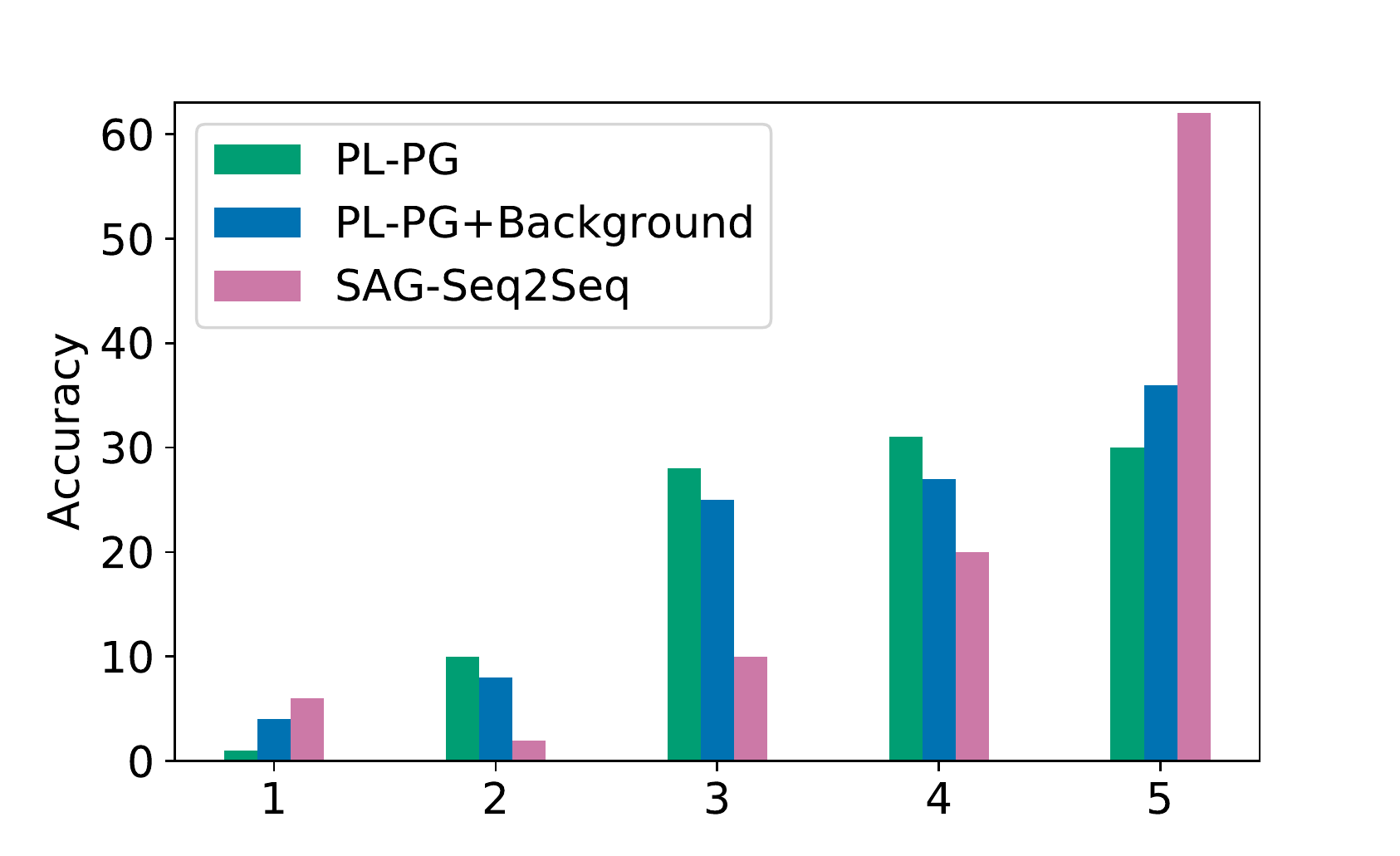}
    \hspace{-5.07mm}
     \includegraphics[width=0.336\textwidth]{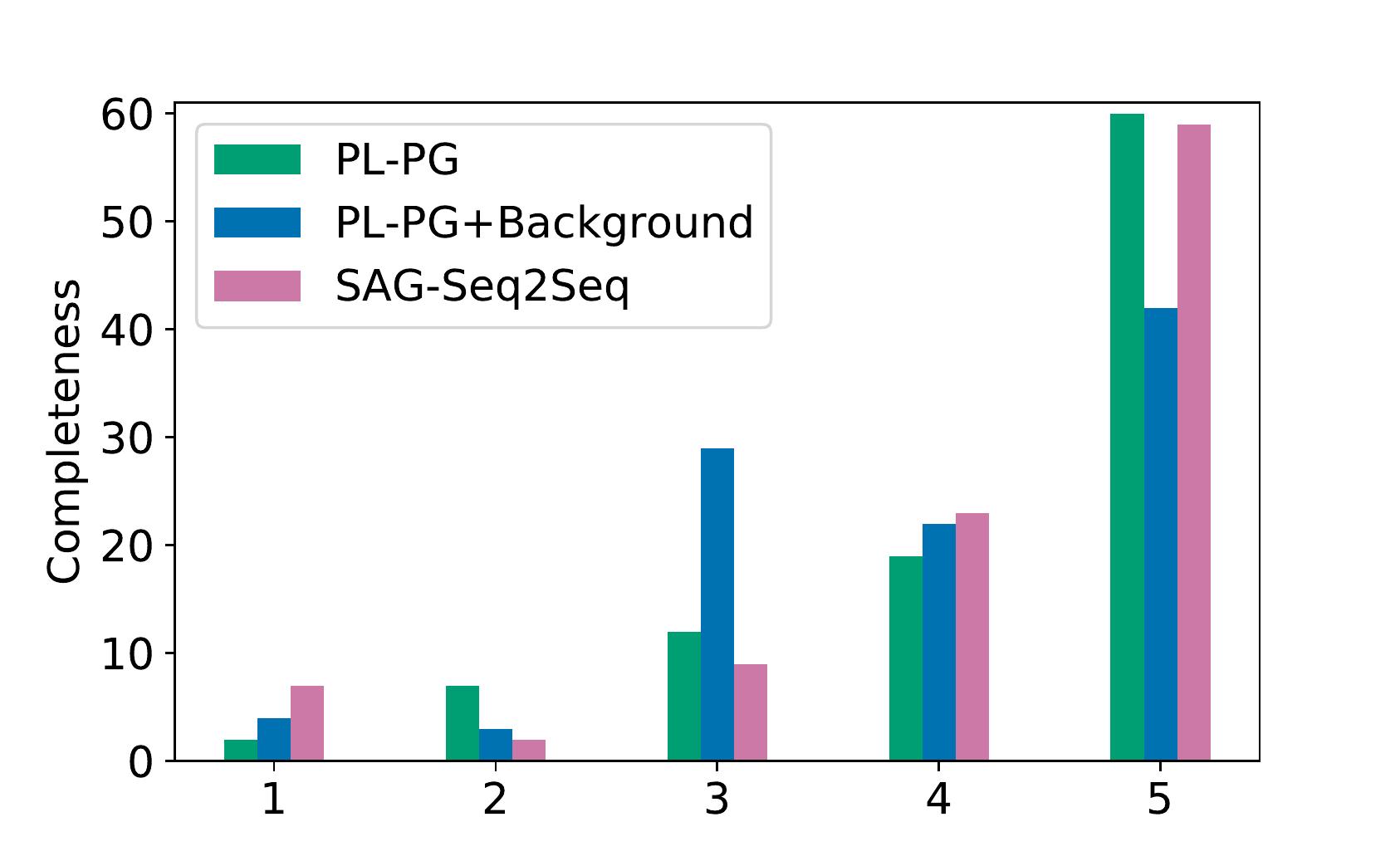}
     \caption{}
   \label{fig:radiologist_results_openi} 
\end{subfigure}

\begin{subfigure}[b]{0.95\textwidth}
    \hspace{-3mm}
    \includegraphics[width=0.38\textwidth]{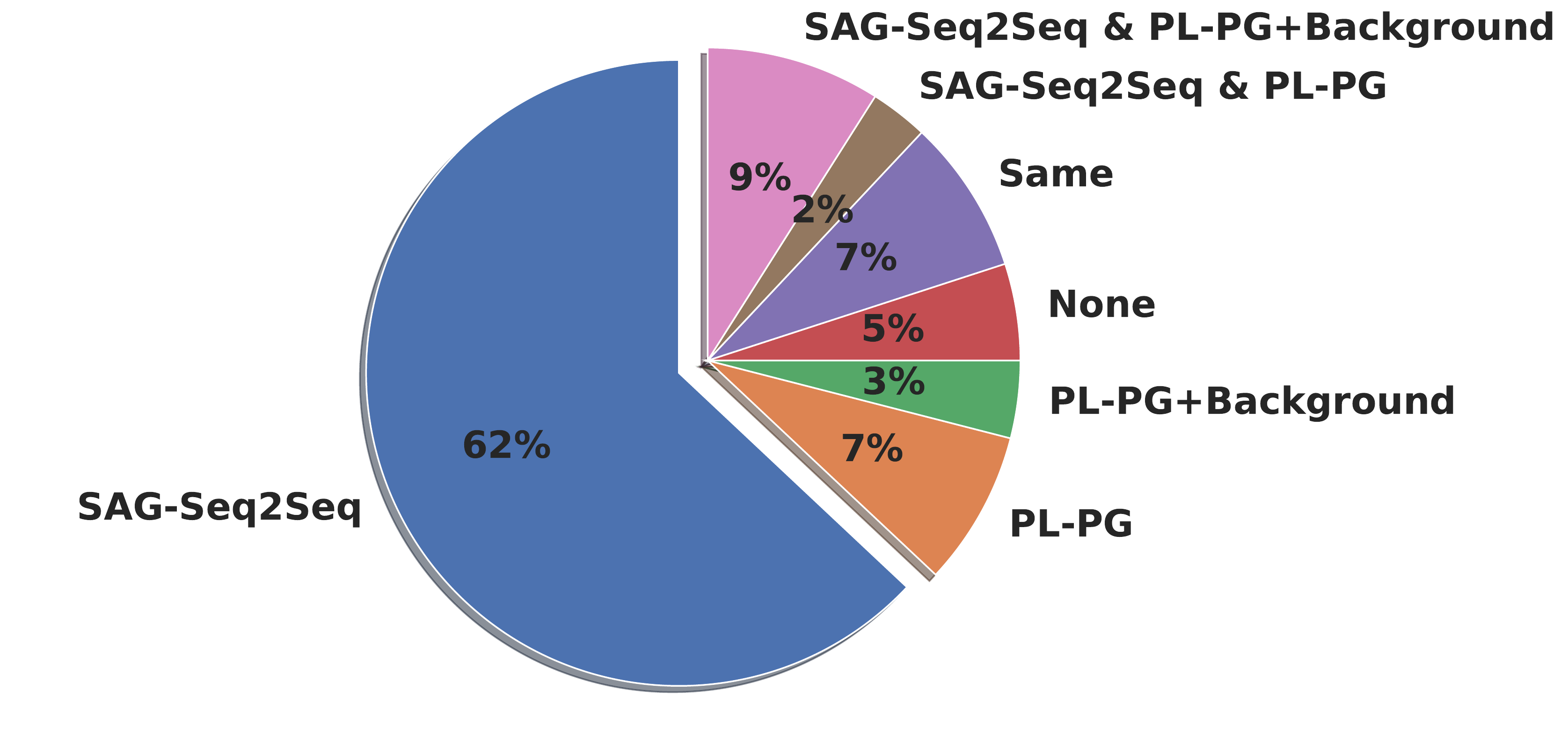}
    \hspace{-1.9mm}
    \includegraphics[width=0.336\textwidth]{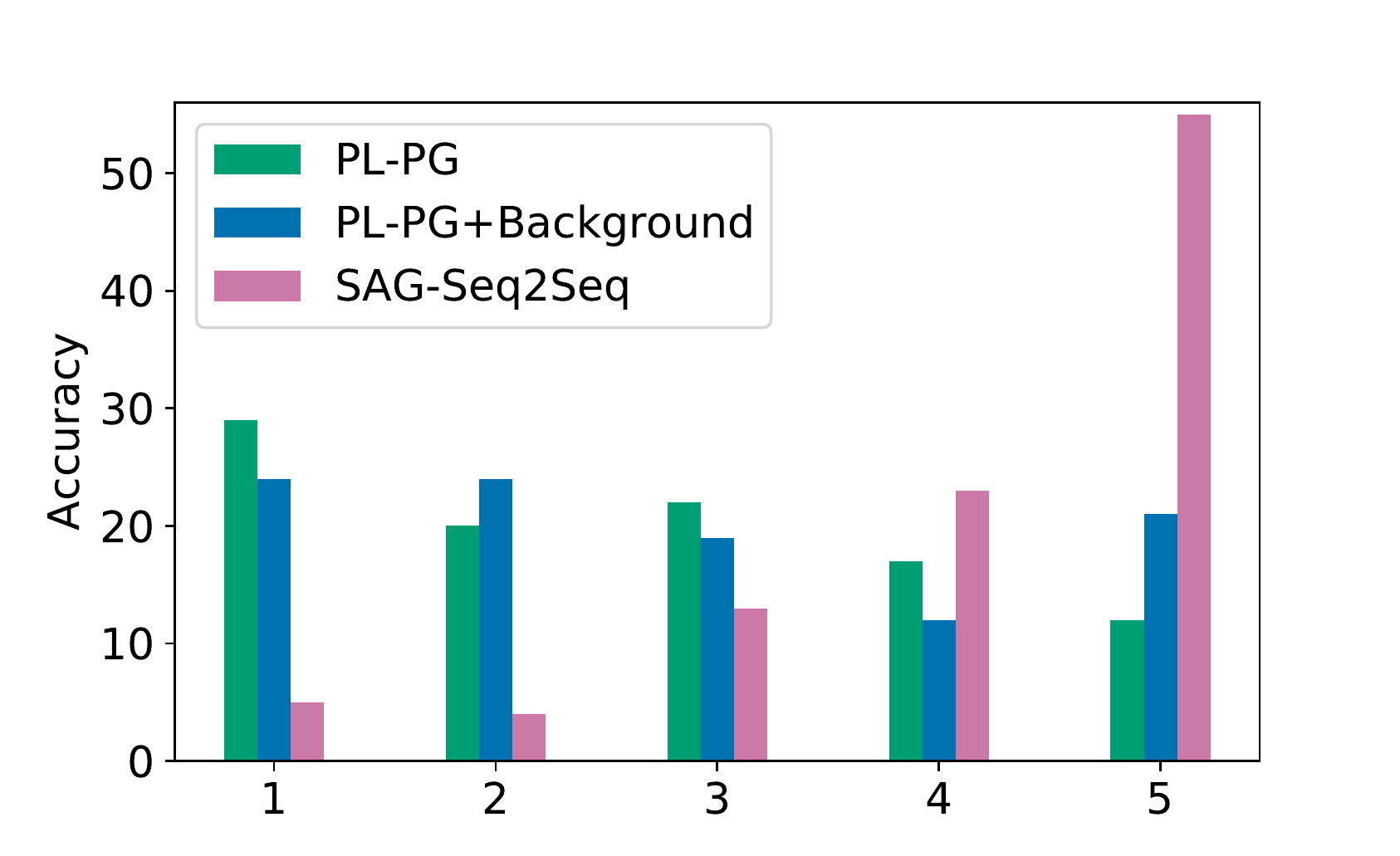}
    \hspace{-5.07mm}
     \includegraphics[width=0.336\textwidth]{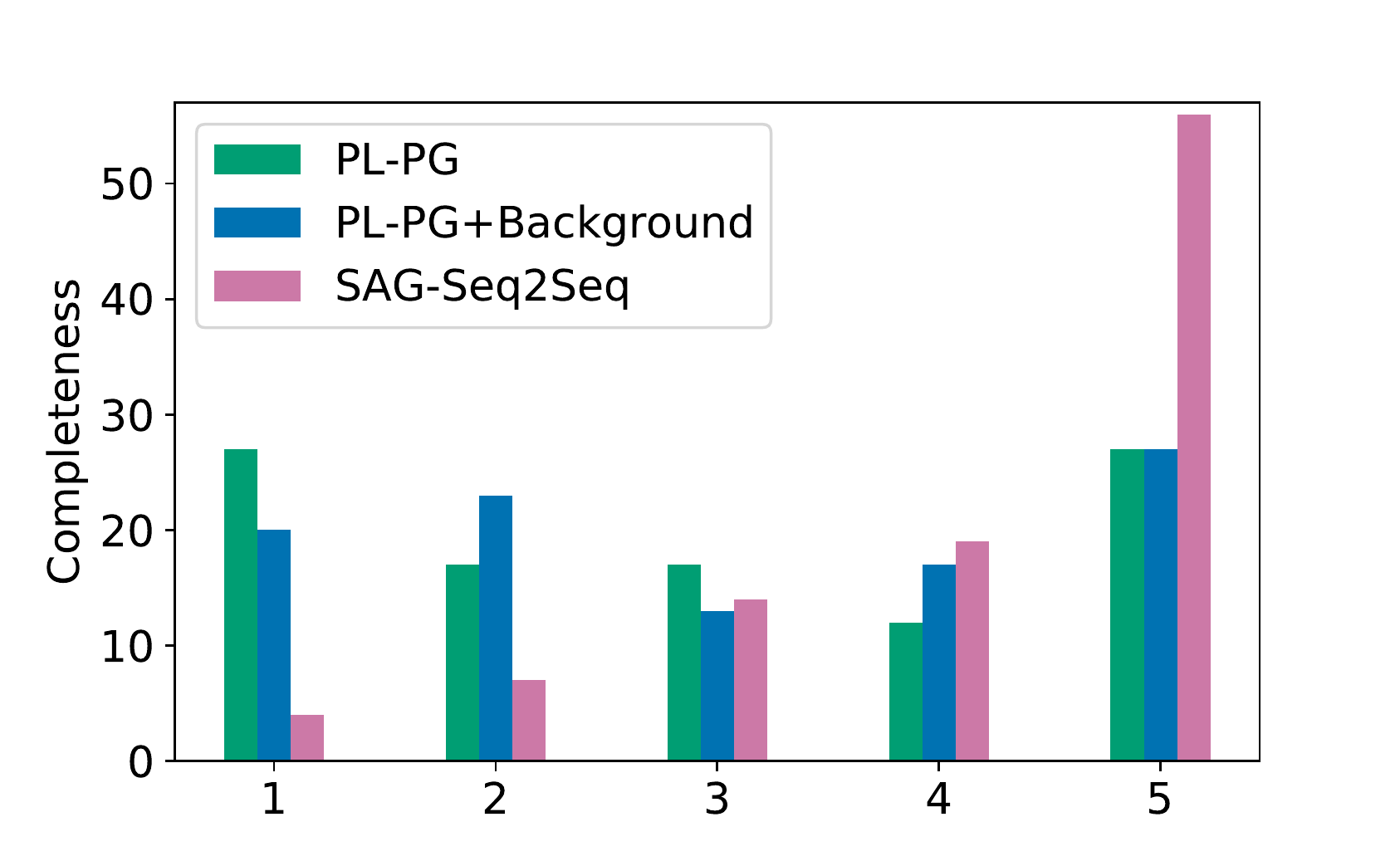}
     \caption{}
   \label{fig:radiologist_results_mimic}
\end{subfigure}
\vspace{-2mm}
\caption{Radiologist results on 100 reports sampled from the (a) OpenI and (b) MIMIC-CXR test sets. The pie charts (left) present the results from each model separately; the center and right histograms show the differences between the three models w.r.t. \textit{Accuracy} and \textit{Completeness}, respectively.}
\vspace{-3mm}
\end{figure}
\vspace*{-1mm}
\subsection{Manual Evaluation by a Radiologist}
\label{sec:radiologist_results}
Some works~\cite{zhang2018learning,macavaney2019ontology} showed that quantitative metrics such as ROUGE are not appropriate for the medical domain where clinical correctness is critical. In addition to automatic evaluation, they used human evaluation. Therefore, we also conducted validity assessments with a radiologist to better understand our system-generated annotations. We ran the top three best performing models (PL-PG, PL-PG+Background, and SAG-Seq2Seq) over one hundred randomly sampled reports from the OpenI test set. Each report, along with the models' generated annotations, is presented to the radiologist. To prevent potential bias, we randomly rearranged the predicted annotations among the models. We asked the radiologist to (1) select the best model from the three models and (2) score the given annotations independently on a scale of 1 to 5 (worst to best) w.r.t. two metrics: \textit{Accuracy} and \textit{Completeness}. The \textit{Accuracy} indicates whether the generated annotations are entirely accurate (score: 5) or contain critical errors (score: 1). The \textit{Completeness} measures whether all important and necessary information is covered (score: 5) or whether key points are missing from the generated annotations (score: 1). The results are presented in Figure~\ref{fig:radiologist_results_openi}.
From the pie chart, we can observe that the radiologist mostly preferred our proposed model. When the two baselines are compared, we can see that the radiologist favors the PL-PG model over the PL-PG+Background model. The two histograms (Figure~\ref{fig:radiologist_results_openi}) indicate that SAG-Seq2Seq achieves 86\% \textit{Accuracy} while both baseline models achieve around 76\%. 
On the other hand, the \textit{Completeness} results are as follows: 85\% for SAG-Seq2Seq, 86\% for PL-PG, and 79\% for PL-PG+Background. In most cases, we discovered that our model generates fewer yet more accurate annotations than the baselines. Overall, the results suggest that our model produces clinically reasonable annotations on the OpenI corpus.\\
\phantom{iiiii}Next, we evaluated our model’s generalizability in generating annotations from an unannotated dataset from another institution. One hundred reports were randomly selected from the MIMIC-CXR dataset for this experiment. To prepare the data for the radiologist, the same procedure from the previous experiment was followed. The results are summarized in Figure~\ref{fig:radiologist_results_mimic}.
In this cross-organizational evaluation, we found that our model considerably outperforms the baseline models and generates clinically reasonable annotations for out-of-sample reports produced at another institution; which can be of benefit to further NLP research with applications in the radiological domain.\\
\phantom{iiiii}In summary, by learning sentence-level representations SAG-Seq2Seq preserves terms relevant to the related annotations. However, its alternatives produce representations on a higher (paragraph) resolution and may miss specific, more granular, annotation-related terms since their influence on the final paragraph representation is diminished by the time the representation is generated.
\vspace*{-4mm}
\section{Conclusion}
\vspace*{-1mm}
\label{sec:conclusion}
This paper studied how to automatically generate report annotations 
to help radiologists reduce the time they devote to annotating reports. We proposed SAG-Seq2Seq, an approach that consists of a rule-SCB-based algorithm to match sentences with annotations, followed by learning a sequence-to-sequence neural model that maps matched sentences to their semi-structured representations. We demonstrate the effectiveness of SAG-Seq2Seq using both quantitative evaluation and qualitative judgment of a radiologist on two publicly available clinical datasets of radiology reports from different providers.

\clearpage

\bibliographystyle{unsrt,splncs04}
%

\clearpage

\appendix

\end{document}